\documentclass[10pt,twocolumn,letterpaper]{article}

\usepackage[pagenumbers]{cvpr} %

\usepackage{booktabs}
\usepackage{multirow}
\usepackage{subcaption}
\usepackage{amsmath,amssymb}
\usepackage{xspace}
\usepackage{stfloats}
\usepackage{float}

\newcommand{\method}{DEA\xspace}
\newcommand{\xres}{EGA\xspace}
\newcommand{\gradcam}{GPA\xspace}
\newcommand{\occlusion}{RIA\xspace}
\newcommand{\methodfull}{Dual-Evidence Attribution\xspace}

\newcommand{\occlusionfull}{Region Intervention Attribution\xspace}

\definecolor{cvprblue}{rgb}{0.21,0.49,0.74}
\usepackage[pagebackref,breaklinks,colorlinks,allcolors=cvprblue]{hyperref}

\def\paperID{25} %
\def\confName{CVPR}
\def\confYear{2026}

\title{Toward Faithful Segmentation Attribution via Benchmarking and Dual-Evidence Fusion}

\author{Abu Noman Md Sakib, OFM Riaz Rahman Aranya, Kevin Desai, Zijie Zhang\\
The University of Texas at San Antonio\\
{\tt\small \{abunomanmd.sakib, ofmriazrahman.aranya, kevin.desai, zijie.zhang\}@utsa.edu}
}

\begin{document}
\maketitle
\begin{abstract}
Attribution maps for semantic segmentation are almost always judged by visual plausibility. Yet looking convincing does not guarantee that the highlighted pixels actually drive the model's prediction, nor that attribution credit stays within the target region. These questions require a dedicated evaluation protocol. We introduce a reproducible benchmark that tests intervention-based faithfulness, off-target leakage, perturbation robustness, and runtime on Pascal VOC and SBD across three pretrained backbones. To further demonstrate the benchmark, we propose \methodfull~(\method), a lightweight correction that fuses gradient evidence with region-level intervention signals through agreement-weighted fusion. \method increases emphasis where both sources agree and retains causal support when gradient responses are unstable. Across all completed runs, \method consistently improves deletion-based faithfulness over gradient-only baselines and preserves strong robustness, at the cost of additional compute from intervention passes. The benchmark exposes a faithfulness--stability tradeoff among attribution families that is entirely hidden under visual evaluation, providing a foundation for principled method selection in segmentation explainability. Code is available at 
\href{https://github.com/anmspro/DEA}{https://github.com/anmspro/DEA}
\end{abstract}

\section{Introduction}
\label{sec:intro}

Post-hoc attribution methods for deep neural networks identify which input regions are causally responsible for a given prediction. In image classification, gradient-based~\cite{zhou2016cam,selvaraju2017grad,chattopadhyay2018gradcampp,wang2020scorecam,desai2020ablationcam,draelos2020hirescam,sundararajan2017axiomatic} and perturbation-based~\cite{zeiler2014visualizing,petsiuk2018rise,fong2017meaningful,fong2019extremal} methods offer well-characterised trade-offs between spatial resolution, computational cost, and faithfulness. Semantic segmentation poses a distinct challenge: an attribution map must explain not only the presence of a target class but whether the model's evidence is correctly localised within a predicted region without assigning importance to spatially disjoint areas. Despite this, segmentation attribution methods remain predominantly gradient-based CAM variants~\cite{zhou2016cam,selvaraju2017grad,vinogradova2020towards,hasany2023segxrescam}, evaluated almost exclusively by visual plausibility. This criterion does not test causal faithfulness. Two questions remain unexamined: does occluding the highest-attributed pixels within the target region reduce the model's confidence in that region, and do attribution maps assign substantial importance to pixels outside the target mask? A method that fails either test may still produce convincing heatmaps, as gradient activations can reflect feature co-occurrence rather than causal evidence~\cite{adebayo2018sanity}.

We introduce a reproducible benchmark for segmentation attribution faithfulness, formalising two evaluation axes absent from prior work: \emph{target deletion faithfulness}, measuring the causal dependence of region-level confidence on highest-attributed pixels, and \emph{absolute off-target leakage}, quantifying attribution credit outside the target mask. Together with perturbation robustness and runtime, these axes enable principled multi-criteria comparison of segmentation attribution methods. We demonstrate the benchmark with \methodfull~(\method), showing that 
it surfaces faithfulness differences invisible to visual inspection. Our contributions are as follows:
\begin{itemize}
    \item A reproducible segmentation attribution benchmark comprising 
    intervention-based faithfulness tests, off-target leakage, 
    perturbation robustness, and runtime profiling across three 
    backbones on Pascal VOC and SBD.
    \item \methodfull~(\method), a lightweight dual-evidence correction 
    fusing gradient and intervention signals, improving deletion 
    faithfulness over gradient baselines.
    \item All per-sample outputs and aggregation scripts are released 
    for independent verification.
\end{itemize}

\section{Related Work}
\label{sec:related}

\noindent\textbf{Gradient-based attribution for dense prediction.}
CAM~\cite{zhou2016cam} and Grad-CAM~\cite{selvaraju2017grad} produce 
class-discriminative maps by weighting activations with globally pooled 
gradients, but discard spatial information through average pooling. 
Grad-CAM++~\cite{chattopadhyay2018gradcampp}, 
Score-CAM~\cite{wang2020scorecam}, 
Ablation-CAM~\cite{desai2020ablationcam}, and 
HiResCAM~\cite{draelos2020hirescam} address different limitations of 
this pooling step. For segmentation, 
Vinogradova~\etal~\cite{vinogradova2020towards} restricted the gradient 
signal to a masked target region (Seg-Grad-CAM), and 
Hasany~\etal~\cite{hasany2023segxrescam} further improved spatial 
specificity with elementwise weighting (Seg-XRes-CAM, our \xres 
baseline). Neither work evaluates causal faithfulness: both validate 
explanations by visual comparison rather than direct intervention tests.

\noindent\textbf{Perturbation and intervention-based attribution.}
Occlusion-based methods~\cite{zeiler2014visualizing} estimate pixel 
importance by masking regions and measuring the change in model output, 
providing a direct causal signal at higher computational cost. 
RISE~\cite{petsiuk2018rise}, meaningful 
perturbations~\cite{fong2017meaningful}, and extremal 
perturbations~\cite{fong2019extremal} extend this with randomised and 
optimised masking schemes. These methods are model-agnostic but their 
evaluation in segmentation remains limited to visual plausibility. Our 
\occlusion baseline and the intervention component of \method adopt the 
same causal viewpoint within the segmentation evaluation loop.

\noindent\textbf{Broader attribution context.}
Beyond CAM-style heatmaps, integrated 
gradients~\cite{sundararajan2017axiomatic}, concept-based 
testing~\cite{kim2018tcav}, and Shapley-value 
approximations~\cite{lundberg2017unified} have shaped widely used 
desiderata for explanation quality, including sensitivity, 
implementation invariance, and completeness. As vision models expanded beyond standard 
CNNs~\cite{dosovitskiy2021vit,woo2023convnextv2,wang2023internimage,jain2023oneformer,kirillov2023segmentanything}, dedicated attention-based and propagation-based explanation methods 
followed~\cite{chefer2021transformer,chefer2021generic,fel2023craft,wang2023botcl,tan2024opencbm,rao2024dncbm,benou2025showandtell}, 
often revealing that attribution behaviour varies substantially across 
architectures. However, the majority of these methods and their 
evaluation protocols target image classification, leaving dense 
prediction tasks without comparable evaluation standards. These lines 
of work motivate the need for faithfulness evaluation that goes beyond 
visual plausibility and accounts for the spatial structure specific to 
segmentation.

\noindent\textbf{Faithfulness evaluation.} Adebayo~\etal~\cite{adebayo2018sanity} showed that many saliency methods produce outputs largely independent of learned weights. Hooker~\etal~\cite{hooker2019benchmark} proposed ROAR, measuring faithfulness by accuracy degradation after retraining on data with top-attributed pixels removed. Yeh~\etal~\cite{yeh2019infidelity} introduced infidelity and sensitivity criteria. These works establish that visual plausibility is insufficient~\cite{bansal2020sam,rao2022betterattr,fel2022stability,behzadikhormouji2023protocol,kim2022hive,kapishnikov2019xrai,srinivas2019fullgrad,chen2018l2x,ribeiro2018anchors,ross2017right,zhao2020counterfactual}, but operate in the classification setting. ROAR requires retraining, and none address off-target leakage specific to dense prediction. Our benchmark instantiates inference-time deletion and leakage tests directly within the segmentation loop, requiring no retraining and explicitly accounting for the spatial structure of dense prediction targets.

\begin{figure}[t]
    \centering
    \includegraphics[width=\linewidth]{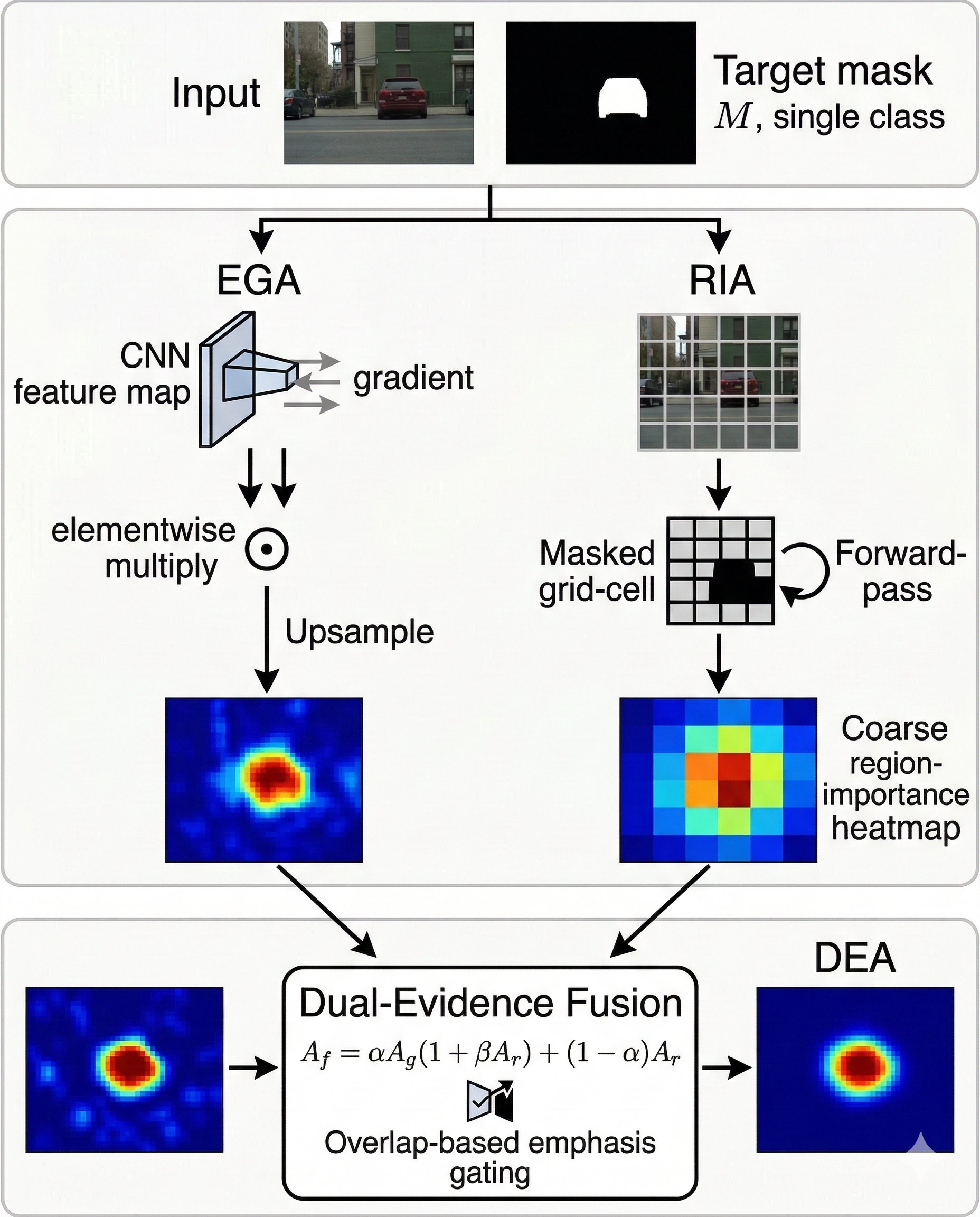}
    \caption{Overview of \method. Elementwise gradient evidence (\xres) and region intervention evidence (\occlusion) are combined through multiplicative agreement and residual intervention support.}
    \label{fig:method_overview}
\end{figure}

\section{Method}
\label{sec:method}
We first define the region-level attribution objective for semantic segmentation, then introduce the dual-evidence correction used in DEA, and finally specify the evaluation metrics. Figure~\ref{fig:method_overview} illustrates the pipeline.

\subsection{Problem Setup}
Given an input image $x \in \mathbb{R}^{3 \times H \times W}$ and a pretrained segmentation model $f$, we study explanations for a target class $c$ and target mask $M \in \{0,1\}^{H \times W}$. Let $z=f(x)$ be per-pixel logits and let $p_c(x)$ denote the softmax probability map for class $c$. We evaluate evidence at the region level through
\begin{equation}
s_c(x,M)=\frac{\sum_{u,v} M_{uv}\,p_c(x)_{uv}}{\sum_{u,v} M_{uv}+\epsilon},
\label{eq:region_score}
\end{equation}
which is the masked mean class probability inside the target region. An attribution method outputs a heatmap $A\in[0,1]^{H\times W}$, and we test whether high-valued pixels in $A$ are causally important for $s_c(x,M)$.

\subsection{Dual-Evidence Attribution}
\label{sec:fusion}
We compare three base attributions that expose complementary behavior. \gradcam uses gradient pooling at the selected feature layer, \xres uses elementwise gradient-activation products at the same layer, and \occlusion computes intervention deltas by masking fixed grid regions and measuring the corresponding drop in \eqref{eq:region_score}. All maps are min-max normalized to $[0,1]$.

Let $A_g$ be the \xres map and $A_r$ the \occlusion map. Our corrected attribution is
\begin{equation}
A_f=\alpha A_g\odot(1+\beta A_r)+(1-\alpha)A_r,
\label{eq:fusion}
\end{equation}
where $(\alpha,\beta)$ control the balance between fine gradient structure and intervention support, and $\odot$ denotes elementwise multiplication. The multiplicative term increases weight on pixels where both sources agree, while the residual $A_r$ term keeps coarse but causally supported evidence when gradient responses are unstable.

\subsection{Metrics}
\label{sec:metrics}
For a heatmap $A$, let $S_t(A,M,k)$ be the top-$k$ fraction of pixels within $M$, and let $S_o(A,M,k)$ be the top-$k$ fraction outside $M$. Using mean-value occlusion operator $\mathcal{O}(x,S)$, we define target deletion drop
\begin{equation}
\mathrm{TDD}=\frac{s_c(x,M)-s_c(\mathcal{O}(x,S_t),M)}{|s_c(x,M)|+\epsilon},
\label{eq:tdd}
\end{equation}
and off-target deletion drop
\begin{equation}
\mathrm{ODD}=\frac{s_c(x,M)-s_c(\mathcal{O}(x,S_o),M)}{|s_c(x,M)|+\epsilon}.
\label{eq:odd}
\end{equation}
Large TDD indicates that attributed target pixels are causally important. Large $|\mathrm{ODD}|$ indicates undesired sensitivity to high-valued pixels outside the target mask. We summarize this tradeoff with
\begin{equation}
\mathrm{LeakAbs}=\frac{|\mathrm{ODD}|}{|\mathrm{TDD}|+\epsilon}.
\label{eq:leakabs}
\end{equation}

We also report target insertion gain by starting from a mean-value baseline image and reinserting $S_t(A,M,k)$, perturbation robustness as the average correlation between original and perturbed heatmaps (noise, brightness, contrast, blur, horizontal flip), and wall-clock runtime per explanation. We report absolute off-target metrics for primary ranking and keep the signed leakage ratio as a diagnostic statistic.

\begin{figure}[t]
    \centering
    \includegraphics[width=\linewidth]{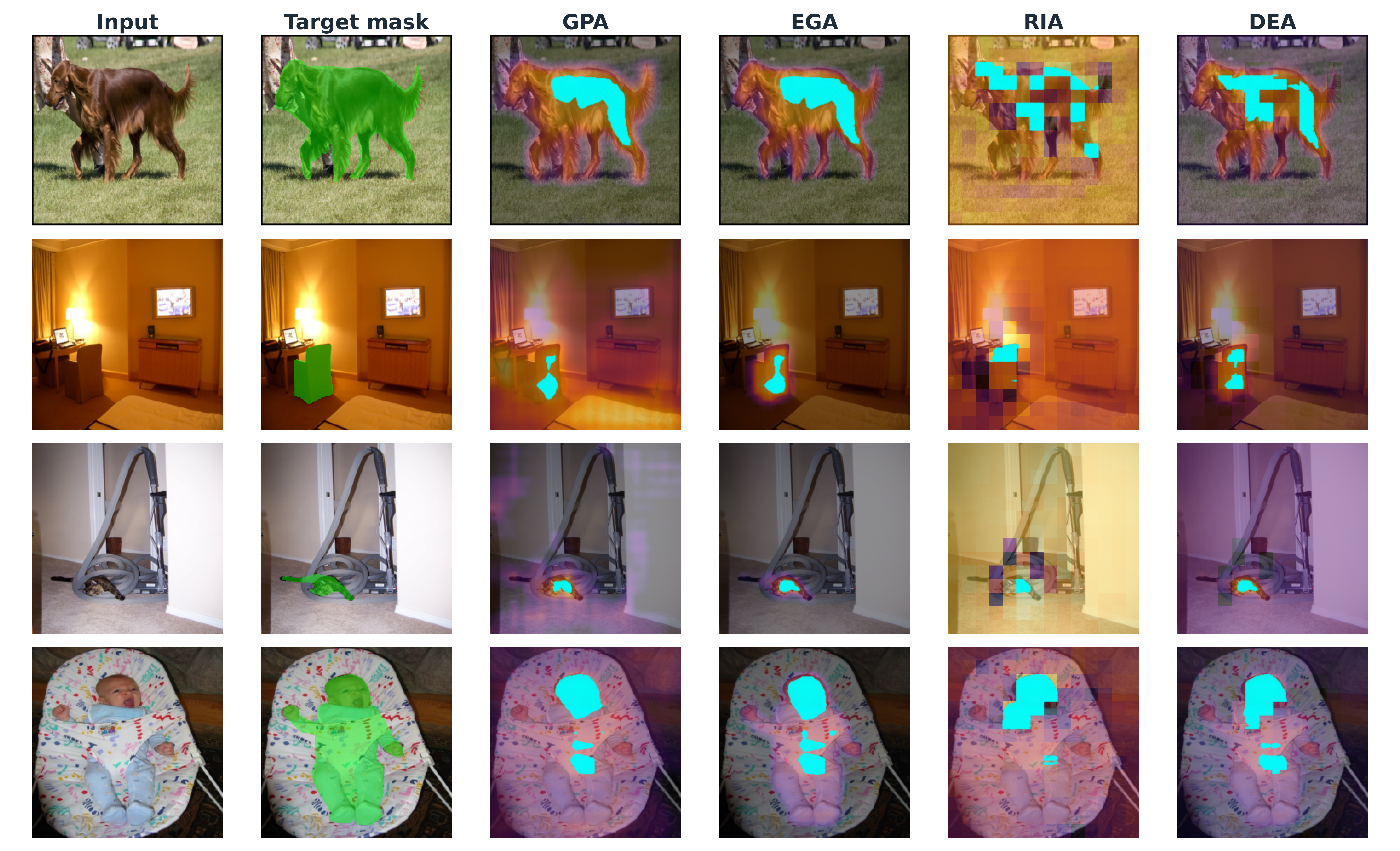}
    \caption{Representative success cases where \method improves target-region faithfulness while preserving spatial focus.}
    \label{fig:qual_main}
\end{figure}

\section{Experimental Setup}
\label{sec:exp_setup}

\subsection{Datasets and Models}
We evaluate on Pascal VOC 2012 and SBD~\cite{everingham2010pascal,hariharan2011semantic}. Images and masks are resized to $224\times224$ and processed with the TorchVision segmentation pipeline. We use pretrained DeepLabV3-ResNet50~\cite{chen2017deeplabv3,chen2018deeplabv3plus}, FCN-ResNet50~\cite{long2015fcn}, and LRASPP-MobileNetV3~\cite{howard2019mobilenetv3}. These choices cover canonical dense-prediction design families used in modern semantic segmentation pipelines~\cite{zhao2017pspnet,zheng2021rethinking,jain2023oneformer,kirillov2023segmentanything}. For each image, we choose a single evaluation target class as the most frequent foreground label in the ground-truth mask (excluding background and ignore label), then define $M$ as that class mask. This protocol avoids degenerate empty targets and makes per-sample comparisons consistent across methods.
\begin{figure*}[t]
    \centering
    \includegraphics[width=\textwidth]{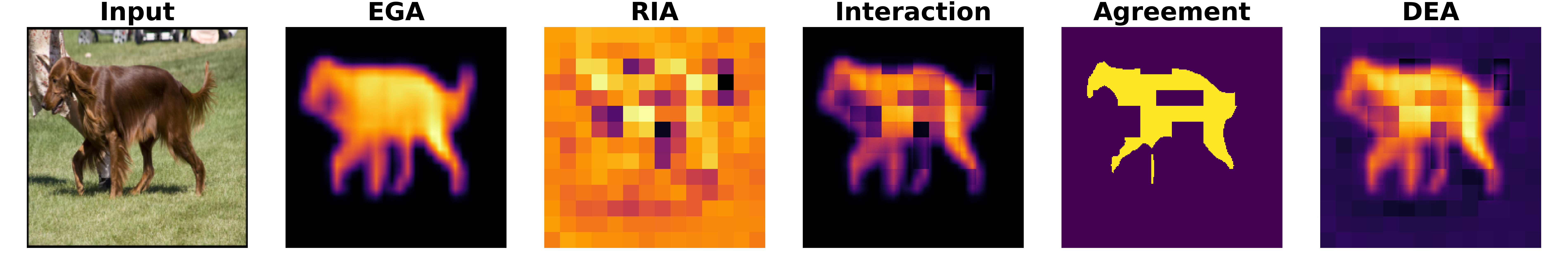}
    \caption{Mechanistic decomposition of \method: elementwise gradient map, region intervention map, interaction, and corrected output (single case).}
    \label{fig:mech}
\end{figure*}

\subsection{Settings}
The compared methods are Gradient-Pooled Attribution 
(GPA, corresponding to Seg-Grad-CAM~\cite{vinogradova2020towards}), 
Elementwise Gradient Attribution 
(EGA, corresponding to Seg-XRes-CAM~\cite{hasany2023segxrescam}), \occlusionfull~(\occlusion), and \method. The top-$k$ fraction used in deletion and insertion tests is $k=0.2$. Region-intervention methods (\occlusion and \method) use grid size 14 by default. For \method, unless noted otherwise, we use $\alpha = 0.65$ and $\beta = 0.35$ from the benchmark implementation. Ablation runs with varied $\alpha$ and $\beta$ confirm that deletion faithfulness remains above \xres for $\alpha \in [0.5, 0.8]$. Robustness is evaluated with perturbation strength $0.03$ over additive noise, brightness shift, contrast change, Gaussian blur, and horizontal flip. Runtime is measured as wall-clock time per explanation call in the same evaluation loop. For SBD, we aggregate all completed runs: six core runs (three backbones, two seeds each), three extra runs, and two DeepLab ablation runs with altered evidence-mixing and grid settings. For VOC, we aggregate six core runs (three backbones, two seeds each). 

\section{Results}
\label{sec:results}

\subsection{Quantitative Results}
\label{sec:quant_results}

Across completed runs, \occlusion reaches the strongest deletion faithfulness and the lowest absolute off-target drop, while \xres remains best in stability and latency. \method lies between these two ends of the tradeoff curve and consistently improves deletion faithfulness over both gradient baselines in every completed run (over \xres: SBD 11/11, VOC 6/6; over \gradcam: SBD 9/9, VOC 6/6). Detailed mean and standard deviation values are provided in the appendix in \cref{sec:appendix_quant,tab:main_results_app}. The tradeoffs are easiest to read relative to \xres, which is the strongest gradient-only baseline in robustness. \method increases target-region deletion faithfulness and reduces off-target absolute drop versus \xres on both datasets, while remaining slower because intervention passes dominate compute. Relative to \occlusion, \method gives up some absolute faithfulness but recovers substantial robustness. This supports a scoped claim: \method is a practical correction when intervention-aligned faithfulness is needed without fully adopting the least stable intervention baseline. Target insertion gain follows a different ordering and tends to favor broader maps, with \gradcam highest on both datasets in our aggregates, and full insertion values are reported in the appendix in \cref{sec:appendix_quant,tab:main_results_app}. We therefore treat insertion as a complementary diagnostic and avoid using it as a single ranking criterion.

\subsection{Qualitative Results}
\label{sec:qualitative}

\cref{fig:qual_main} shows the qualitative behavior behind the aggregate metrics and makes the deletion and leakage tradeoff visually explicit. Across success cases, \method suppresses broad low-confidence context activation that appears in gradient-only maps, while preserving contiguous object structure inside the target region. Compared with \occlusion, the corrected map typically keeps sharper intra-object detail and avoids the block-like over-smoothing introduced by coarse intervention regions. \cref{fig:mech} explains this behavior at the mechanism level. The interaction term acts as a gate that promotes pixels supported by both evidence streams and attenuates pixels favored by only one stream. In the selected mechanistic cases generated by our pipeline, agreement mass is concentrated on target pixels with near-zero off-target overlap, and the final map is consistently tighter than the raw intervention map while remaining less noisy than the gradient map. Additional failure-mode analysis, including thin-boundary under-coverage and clutter-driven residual leakage, is provided in the appendix in \cref{sec:appendix_qual,fig:sbd_failure_app}.

\section{Discussion}
\label{sec:framing}

The central finding is not that \method dominates every axis; it does not. The robust conclusion is that \method reliably shifts gradient-based attribution toward intervention-validated faithfulness, at the predictable cost of intervention-level compute. This is useful in evaluation-heavy settings where explanation quality is more important than millisecond latency, and less attractive for strict real-time constraints where \xres is still preferable. Two limitations remain important. First, our uncertainty reporting currently uses run-level variability, not formal hypothesis tests, so claims should be interpreted as strong empirical trends rather than definitive significance statements. Second, intervention maps are built from fixed grids, which can miss thin structures and contribute to residual leakage in difficult scenes.

\section{Conclusion}
\label{sec:conclusion}

We introduced a reproducible benchmark for segmentation attribution and a dual-evidence correction that combines high-resolution gradient structure with intervention support. Across completed SBD and VOC runs, \method consistently improves deletion-based faithfulness over gradient baselines while preserving high robustness, though it remains much slower than pure gradient methods because intervention passes dominate compute. The empirical picture is intentionally scoped. \method is best viewed as a correction that moves gradient attribution toward intervention-aligned behavior, not as a universal best method across all axes. In our aggregates, pure intervention attribution still leads on absolute faithfulness metrics, while \xres remains strongest on speed and stability. Future work should add formal significance testing, denser and adaptive intervention schemes for boundary-sensitive regions, and faster region-level evaluation so intervention-grounded attribution becomes practical in larger-scale and lower-latency settings.

{
    \small
    \bibliographystyle{ieeenat_fullname}
    \bibliography{main}
}

\appendix
\clearpage
\section{Appendix}
\label{sec:appendix}

\subsection{Additional Qualitative Cases}
\label{sec:appendix_qual}

\begin{figure}[H]
    \centering
    \includegraphics[width=\linewidth]{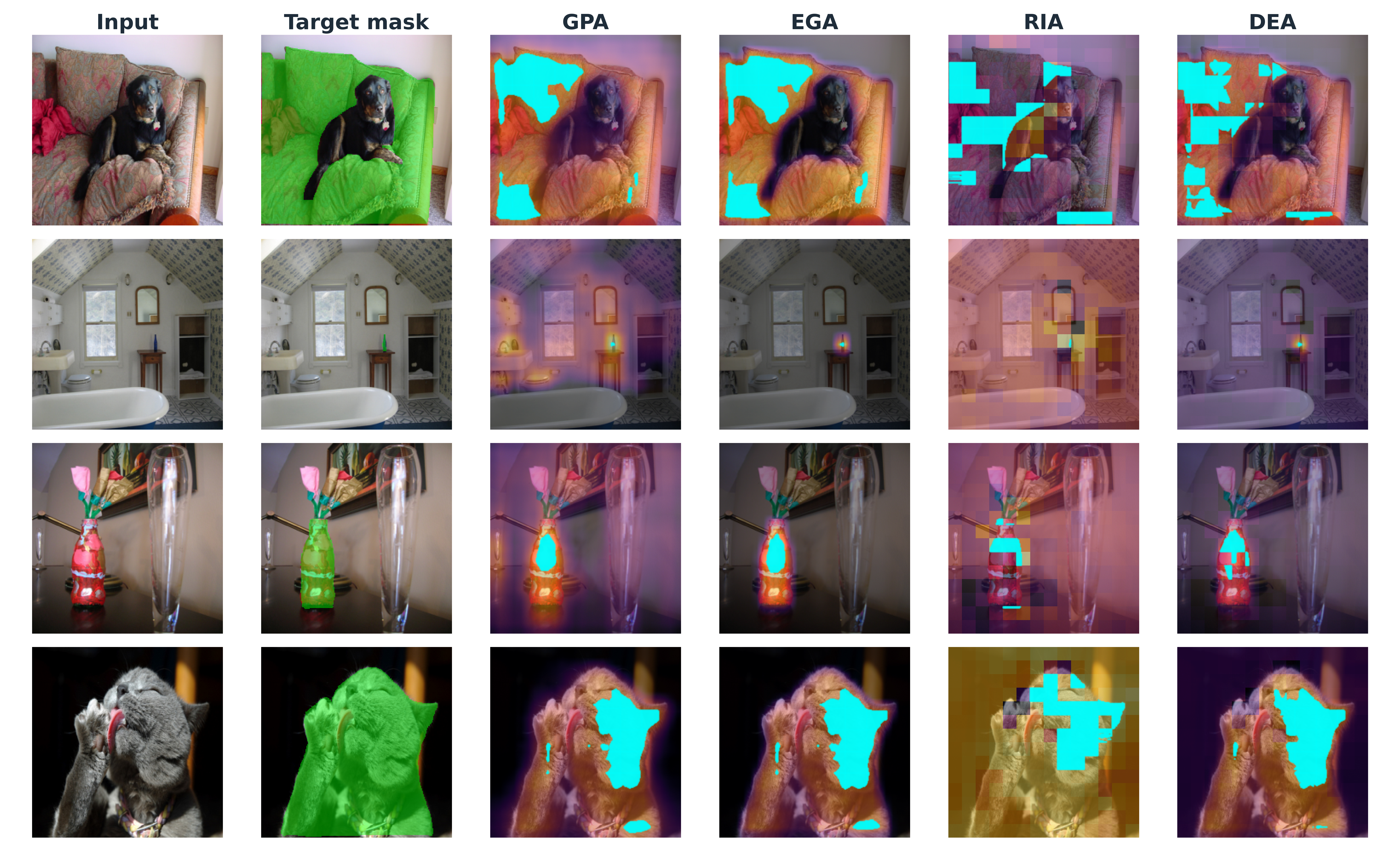}
    \caption{Representative SBD failure cases under the same comparison pipeline used for main-text figures. Residual off-target activation remains in cluttered contexts, and fine boundary detail can be missed on thin target structures.}
    \label{fig:sbd_failure_app}
\end{figure}

Failure cases are shown in Figure~\ref{fig:sbd_failure_app}. Two 
recurring patterns appear. First, thin structures and weak boundaries 
produce under-coverage even when the target object is partially 
highlighted. Second, highly textured co-occurring regions can attract 
non-trivial activation when repeatedly reinforced by intervention 
responses. These failure modes explain the residual off-target 
emphasis in difficult scenes and motivate future work on adaptive 
region partitioning.

\subsection{Quantitative Details}
\label{sec:appendix_quant}

Table~\ref{tab:main_results_app} reports method performance as mean 
and standard deviation across run-level means, so each completed run 
contributes equally to the aggregate (11 SBD runs, 6 VOC runs). The 
insertion-gain column clarifies the complementary behaviour noted in 
the main text: broader maps score higher on insertion even when less 
selective under deletion and leakage diagnostics.

The VOC \xres outlier in off-target absolute drop 
($0.982 \pm 1.228$) is driven by the DeepLabV3 backbone in both 
seeds, where per-run values reach approximately $2.565$, while FCN 
and LRASPP runs remain much lower (${\sim}0.265$ and ${\sim}0.116$ 
respectively). This reflects a backbone-specific concentration effect 
rather than a single-seed anomaly.

\begin{table}[H]
\centering
\setlength{\tabcolsep}{3pt}
\small
\caption{Completed aggregates reported as mean $\pm$ std across runs. Higher is better for TDD, insertion gain, and stability; lower is better for off-target absolute drop.}
\label{tab:main_results_app}
\begin{tabular}{llcccc}
\toprule
Data & Method & TDD $\uparrow$ & OT abs $\downarrow$ & Ins. $\uparrow$ & Stab. $\uparrow$ \\
\midrule
\multirow{4}{*}{\rotatebox{90}{SBD}}
& \gradcam & $.259{\scriptstyle\pm.021}$ & $.748{\scriptstyle\pm.736}$ & $.281{\scriptstyle\pm.196}$ & $.883{\scriptstyle\pm.006}$ \\
& \xres & $.223{\scriptstyle\pm.027}$ & $.268{\scriptstyle\pm.302}$ & $.226{\scriptstyle\pm.199}$ & $.978{\scriptstyle\pm.004}$ \\
& \occlusion & $.453{\scriptstyle\pm.021}$ & $.177{\scriptstyle\pm.125}$ & $.097{\scriptstyle\pm.037}$ & $.827{\scriptstyle\pm.010}$ \\
& \method & $.381{\scriptstyle\pm.030}$ & $.235{\scriptstyle\pm.134}$ & $.114{\scriptstyle\pm.047}$ & $.959{\scriptstyle\pm.009}$ \\
\midrule
\multirow{4}{*}{\rotatebox{90}{VOC}}
& \gradcam & $.270{\scriptstyle\pm.044}$ & $.281{\scriptstyle\pm.133}$ & $.123{\scriptstyle\pm.094}$ & $.894{\scriptstyle\pm.004}$ \\
& \xres & $.287{\scriptstyle\pm.048}$ & $.982{\scriptstyle\pm1.23}$ & $.060{\scriptstyle\pm.009}$ & $.978{\scriptstyle\pm.003}$ \\
& \occlusion & $.522{\scriptstyle\pm.050}$ & $.249{\scriptstyle\pm.184}$ & $.031{\scriptstyle\pm.013}$ & $.821{\scriptstyle\pm.001}$ \\
& \method & $.449{\scriptstyle\pm.043}$ & $.398{\scriptstyle\pm.313}$ & $.074{\scriptstyle\pm.017}$ & $.961{\scriptstyle\pm.004}$ \\
\bottomrule
\end{tabular}
\end{table}

\end{document}